\documentclass[10pt,twocolumn,letterpaper]{article}

\usepackage{wacv}              %

\usepackage{graphicx}
\usepackage{amsmath}
\usepackage{amssymb}
\usepackage{booktabs}

\usepackage{pifont}
\newcommand{\cmark}{\ding{51}}%
\newcommand{\xmark}{\ding{55}}%

\usepackage[pagebackref,breaklinks,colorlinks]{hyperref}

\usepackage[capitalize]{cleveref}
\crefname{section}{Sec.}{Secs.}
\Crefname{section}{Section}{Sections}
\Crefname{table}{Table}{Tables}
\crefname{table}{Tab.}{Tabs.}

\begin{document}

\title{Map-Free Trajectory Prediction with Map Distillation \\ and Hierarchical Encoding}

\author{Xiaodong Liu \quad Yucheng Xing \quad Xin Wang\\
Stony Brook University \\
{\tt\small \{xiaodong.liu, yucheng.xing, x.wang\}@stonybrook.edu}
}

\maketitle

\begin{abstract}

Reliable motion forecasting of surrounding agents is essential for ensuring the safe operation of autonomous vehicles. Many existing trajectory prediction methods rely heavily on high-definition (HD) maps as strong driving priors. However, the availability and accuracy of these priors are not guaranteed due to substantial costs to build, localization errors of vehicles, or ongoing road constructions. In this paper, we introduce \textbf{MFTP}, a \textbf{M}ap-\textbf{F}ree \textbf{T}rajectory \textbf{P}rediction method that offers several advantages. First, it eliminates the need for HD maps during inference while still benefiting from map priors during training via knowledge distillation. Second, we present a novel hierarchical encoder that effectively extracts spatial-temporal agent features and aggregates them into multiple trajectory queries. Additionally, we introduce an iterative decoder that sequentially decodes trajectory queries to generate the final predictions. Extensive experiments show that our approach achieves state-of-the-art performance on the Argoverse dataset under the map-free setting.

\end{abstract}
    
\section{Introduction}
\label{sec:intro}

Trajectory prediction plays a pivotal role in autonomous driving systems to ensure safety. It has drawn increasing attention in the community because of the rapid progress in autonomous driving~\cite{hivt,mtr,qcnet,densetnt,iccv2023adapt,motiondiffuser,seff2023motionlm,iccv2023eigentrajectory,iccv2023forecast-mae,ipcc,fend,bahari2022vehicle,choi2022hierarchical,kuang2020multi,wang2022ptseformer} and publicly available datasets~\cite{argoverse1, argoverse2, waymo}. With the trajectory prediction,  future trajectories of interested agents (\eg, vehicles, pedestrians) are predicted based on their historical trajectories and context information, such as HD maps and traffic signals. This task remains challenging because of diversified driving behaviors, the complexity of the real-world environment, and abrupt events.

\begin{figure}[t]
  \centering
  \begin{subfigure}{0.9\linewidth}
    \includegraphics[width=0.99\textwidth]{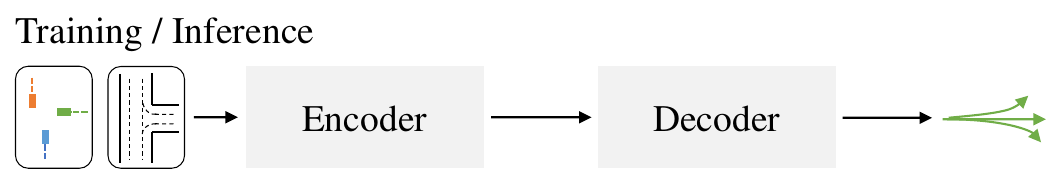}
    \caption{Map-based Methods}
    \label{fig:map-based}
  \end{subfigure}
  \vfill
  \begin{subfigure}{0.9\linewidth}
    \includegraphics[width=0.99\textwidth]{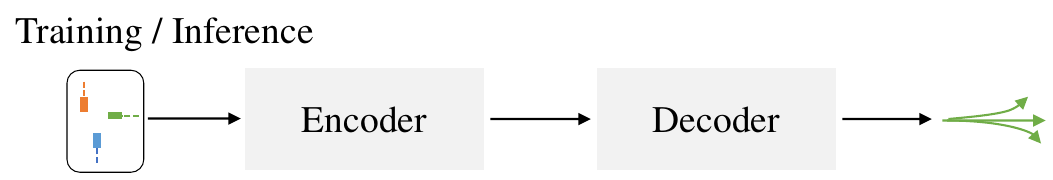}
    \caption{Map-free Methods}
    \label{fig:map-free}
  \end{subfigure}
  \vfill
  \begin{subfigure}{0.9\linewidth}
    \includegraphics[width=0.99\textwidth]{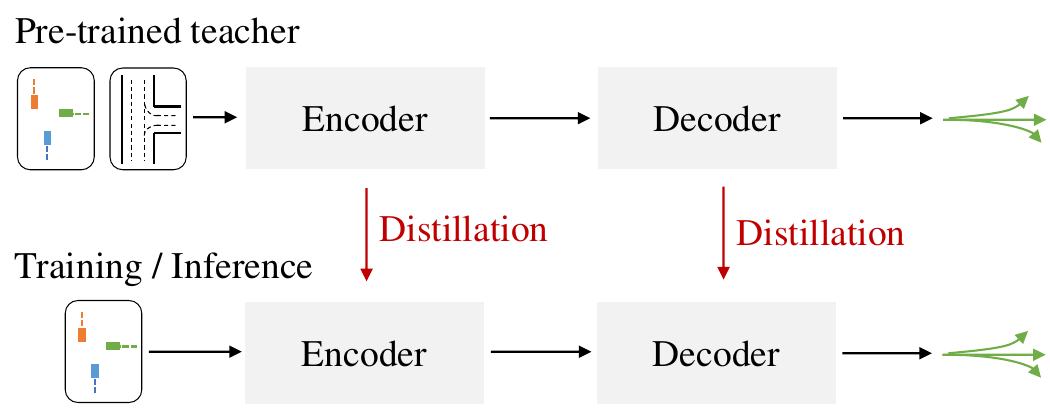}
    \caption{Ours}
    \label{fig:map-free-ours}
  \end{subfigure}
  \caption{\textbf{The differences between existing methods with ours.} Existing map-based methods utilize map information during both training and inference, whereas map-free methods do not. In contrast, our method employs a pre-trained map-based teacher network to distill map priors into a map-free student network.}
  \label{fig:method-comparison}
\end{figure}

Existing methods for trajectory prediction generally fall into two categories: map-based and map-free methods. Map-based methods treat trajectory prediction as a multimodal problem, with HD maps serving as a crucial modality for interactions with agents. These methods focus on optimizing the representation of contextual information, with the maps acting as strong driving priors to influence the future behaviors of agents. For example, LAFormer~\cite{laformer} incorporates a dense lane-aware estimation module to select the most possible lanes for future trajectories. MTR~\cite{mtr} encodes the interaction between agents and map via a transformer encoder and dynamically selects the closest lanes on the map for possible destinations during the trajectory generation. 
Wayformer~\cite{wayformer} explores various fusion strategies for multimodal inputs and evaluates the effectiveness of different encoder architectures for context modeling. While these map-based methods have achieved impressive results on publicly available datasets, they face limitations due to the limited accessibility and occasional inaccuracies of map data in practical driving scenarios. These limitations raise security concerns and underscore the need for map-free trajectory prediction methods.

However, map-free trajectory prediction, while essential for real autonomous driving systems, has received comparatively less attention than its map-based counterparts. Only recently have researchers started to recognize the significance of map-free trajectory prediction in autonomous driving and have attempted to address the prediction problem without relying on HD maps. For example, CRAT-Pred~\cite{cartpred} applies a graph convolutional neural network to model social interactions among vehicles based solely on historical data, predicting future trajectories without utilizing maps. Xiang \etal~\cite{fastmapfree} proposes a two-stage framework to extract agents' spatiotemporal interactions using LSTM and transformer, also predicting the future trajectories without map information. Despite the notable progress made by these map-free methods, their performance still lags behind map-based methods, primarily due to the absence of map priors. A natural question arises: \textit{Can map-free trajectory prediction also benefits from map priors like map-based methods?}

In this paper, we propose MFTP, a trajectory prediction method that utilizes map priors during the training without sacrificing the model's ability for map-free prediction (see \cref{fig:method-comparison}). Specifically, we achieve this by distilling map priors from a pre-trained map-based teacher network into a map-free student, which allows leveraging map priors to facilitate motion forecasting while remaining proficient in predicting future trajectories without maps. Additionally, we propose a novel hierarchical encoder to aggregate multiple levels of spatial-temporal features of agents into hierarchical queries. These queries are then fused into several trajectory queries for each agent. Subsequently, these fused trajectory queries are fed into our iterative decoder to sequentially generate future trajectories.

Our experiments demonstrate that our method achieves a new state-of-the-art performance on the Argoverse~\cite{argoverse1} dataset under the map-free setting. We hope our work can facilitate and draw more attention to map-free trajectory prediction. Our primary contributions can be summarized as follows:
\begin{enumerate}
  \item We propose a map-free trajectory prediction method that leverages map priors during the training while preserving the ability for map-free prediction during the inference.
  \item We introduce a hierarchical encoder that extracts multi-level spatial-temporal features of agents and squeezes them into multiple trajectory queries. These queries are subsequently employed in generating future trajectories using our iterative decoder.
  \item We conduct extensive experiments, and our results show the superior performance of our method on the Argoverse dataset under the map-free setting.
\end{enumerate}

\section{Related Work}
\label{sec:related-work}

\paragraph{Trajectory prediction for autonomous driving.}
Trajectory prediction has witnessed significant advancements in recent years, driven by the rapid progress in autonomous driving, the availability of large-scale public datasets~\cite{argoverse1,argoverse2,waymo,caesar2020nuscenes}, and associated challenges. Most of the existing methods treat maps as strong driving priors and focus on improving the representation of scene elements and modeling their interactions.

VectorNet~\cite{gao2020vectornet} abandons rasterized representation, which organizes the map and trajectories in an image-like format, and introduces a vectorized representation to the field, greatly simplifying the input representation and reducing computation costs. Many subsequent methods exploit this representation and encode the scene elements with advanced structures, such as transformer~\cite{mtr,mtr++,qcnet,iccv2023adapt,li2022sit}, graph neural networks (GNNs)~\cite{cartpred,fastmapfree}, \etc. To better model interactions, HiVT~\cite{hivt} proposes a hierarchical framework to first extract local relationships and then capture global interactions. MTR++~\cite{mtr++} proposes a symmetric context modeling for better scene representation and simultaneous prediction of the multimodal motion for multiple agents. QCNet~\cite{qcnet} employs a query-centric reference frame to represent the context and leverages the relative information, such as position, heading, and velocity, to extract the relationship between elements. Nevertheless, these methods rely heavily on the map information, leading to significant performance degradation during the inference when the map is absent.

In contrast, map-free methods aim to tackle trajectory prediction without relying on maps. CRAT-Pred~\cite{cartpred} exclusively utilizes the information about agents for trajectory prediction. It employs LSTM to encode temporal features of vehicles, and incorporates GNNs and attention mechanisms to model interactions between vehicles. Xiang \etal~\cite{fastmapfree} adopt a two-stage framework, encoding individual agents with LSTM and then modeling interactions between multiple agents using GNN and transformer. However, map-free methods often fail to achieve comparable performance with map-based methods due to the lack of map priors during training. To mitigate this, FOKD~\cite{wang2023enhancing} introduces a feature and output distillation framework aimed at enhancing the performance of map-free variants of existing map-based methods. Nonetheless, as these networks are originally designed for map-based settings, they might not be optimal for map-free prediction. To address this, we propose a novel structure tailored for map-free trajectory prediction that includes relative inputs, a hierarchical encoder, and an iterative decoder, with knowledge distillation serving as an additional booster. Unlike FOKD, we focus on distilling knowledge from trajectory queries to enable the map-free student to learn the priors captured by the teacher.

\begin{figure*}[t]
  \centering
  \includegraphics[width=0.99\textwidth]{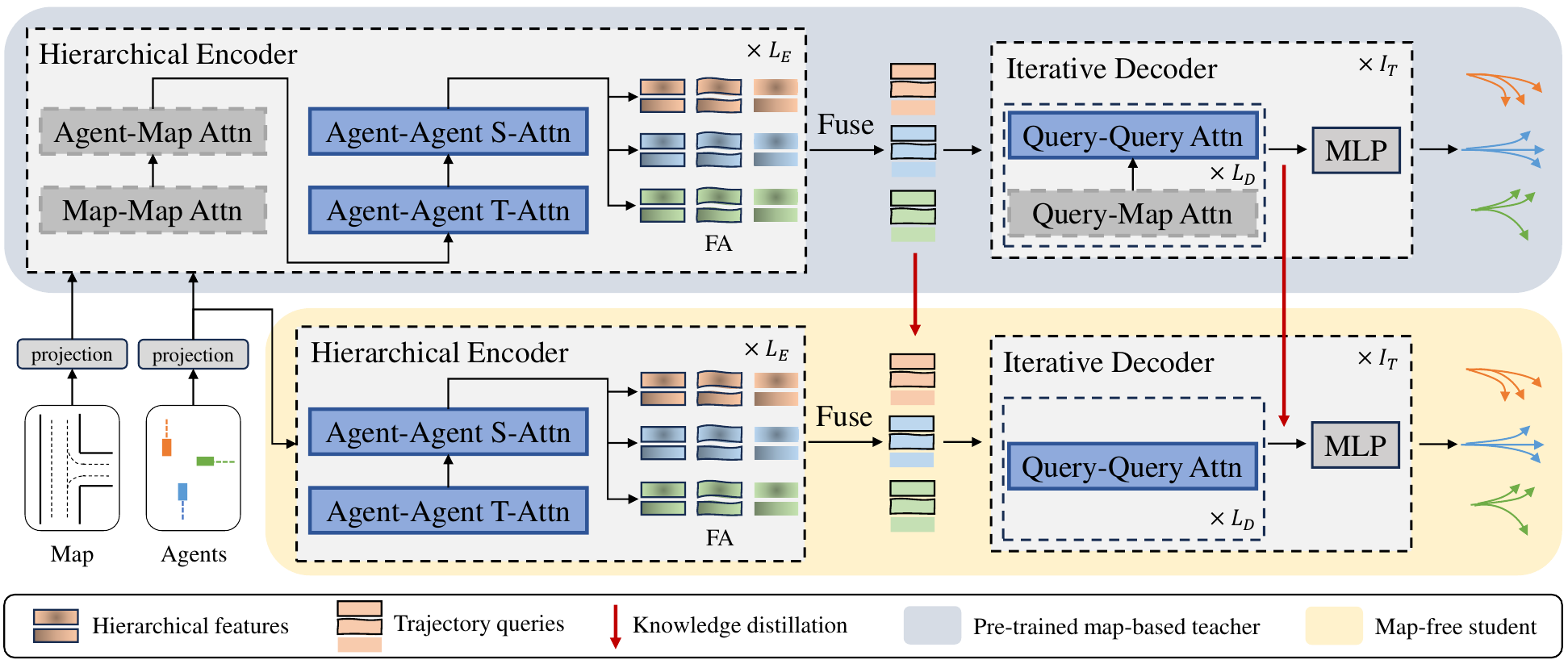}

   \caption{\textbf{Overall framework of MFTP.} MFTP has a pre-trained map-based teacher model and a map-free student model. The student has the same architecture as the teacher except for the map-related modules. The hierarchical agent features are progressively extracted after agent-agent temporal and spatial attention through the Feature Aggregation (FA) module in the encoder, and these features are then fused to form $K$ trajectory queries, corresponding to $K$ multimodal future trajectories. In the teacher network, the agents learn map priors through the agent-map attention module in the encoder stage, and query-map attention module during the decoder stage. Through knowledge distillation of intermediate features, we squeeze map priors into the map-free student network.}
   \label{fig:framework}
\end{figure*}

\paragraph{Knowledge distillation.}
Knowledge distillation involves the transfer of knowledge from a larger network to a smaller one. The concept of knowledge distillation was originally introduced by Hinton \etal~\cite{hinton2015distilling}, and was later used in many tasks, including object detection~\cite{wang2019distilling}, classification~\cite{xu2020feature}, pose estimation~\cite{li2021online}, \etc. According to the distilled knowledge, subsequent works fall into two categories, logits distillation~\cite{hinton2015distilling, zhao2022decoupled, mirzadeh2020improved, guo2020online} and intermediate features distillation~\cite{heo2019comprehensive,heo2019knowledge,kim2018paraphrasing,park2019relational,tian2019contrastive}. Depending on whether the teacher network is trained or not during the distillation process, these methods can also be divided into offline~\cite{li2022shadow,hinton2015distilling,park2019relational,tian2019contrastive} and online distillation~\cite{guo2020online,chen2020online,wu2021peer,qian2022switchable,li2022distilling}. We opt for offline distillation for its simplicity and leverage the knowledge transfer from intermediate features, which is more suitable for enhancing trajectory prediction.

\section{Methodology}
\label{sec:method}

\subsection{Overall Framework}
\label{subsec:overall-framework}
The overall framework of the proposed MFTP is illustrated in ~\cref{fig:framework}.  It comprises two main components: a pre-trained map-based teacher and a distilled map-free student. 
The teacher network takes both map and agents' historical trajectories as inputs, exploiting transformer attention to model map-map, agent-map, and agent-agent interactions at the encoding stage. Simultaneously, multiple distinct hierarchical queries are employed to progressively gather hierarchical spatial and temporal features of agents during the context encoding. Subsequently, these hierarchical features are concatenated and compressed into $K$ trajectory queries via MLPs. These queries serve as the starting points for predicting $K$ multimodal future trajectories for each agent, incorporating not only agent features but also map priors. An iterative decoder is then employed to sequentially decode the future trajectories of agents. With the assistance of the query-map attention module, the network leverages map priors for trajectory decoding. The decoder predicts the trajectories for a given time period in each time step, finishing the prediction after $I_T$ steps. Compared to methods that predict all future points in a single pass or autoregressive decoding methods that output a single point at a time,  our method ensures both prediction accuracy and computational efficiency. The student network follows the same pipeline as the teacher network, except for the map-related modules. We apply Knowledge distillation to trajectory queries after the encoder and to query features before the MLP in the decoder, distilling map priors into our map-free student network.

\subsection{Hierarchical Encoder}
\label{subsec:encoder}
\paragraph{Input representation.}
With vector representation, historical trajectories of agents and map polylines are represented as a set of points in 2D or 3D space. Instead of using the absolute positions of agent trajectories and map points, we opt for relative motion vectors as inputs. Specifically, given an agent's historical trajectory points $\{P_i^t=(x_i^t,y_i^t)|i=1,2,...,N_A, t=-T+1,...,0\}$, where $N_A$ is the number of agents and $T$ is the length of historical points, we calculate motion vectors $\{V_i^t = P_i^t - P_i^{t-1} | t=-T+2,...,0\}$ and the corresponding mask $\{M_i^t = m_i^t \& m_i^{t-1} | t=-T+2,...,0\}$, where the mask is used to select valid features in the following modules. A mask $M_i^t$ is valid only if the historical positions are valid at the time steps $t$ and $t-1$. The same method is applied to process map polylines, yielding displacement vectors between static lane points and corresponding masks. Subsequently, we project these motion vectors to feature space with MLPs and get agent inputs $F_A \in \mathbb{R}^{N_A \times T \times D}$ and map inputs $F_M \in \mathbb{R}^{N_M \times L \times D}$, where D is the feature dimension, $N_M$ is the map polylines and $L$ is the number of points per map polyline. To simplify the map input and reduce computation cost, we squeeze map feature to $F_M \in \mathbb{R}^{N_M \times D}$ via max operation like~\cite{mtr}, representing features for all map polylines.

\begin{figure}[t]
  \centering
   \includegraphics[width=0.43\textwidth]{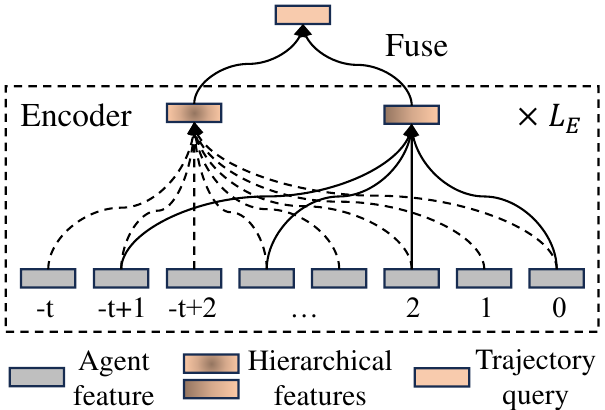}

   \caption{\textbf{Illustration of hierarchical feature aggregation and fusion.} When provided with multiple features of a single agent, our approach employs multiple queries to extract different levels of features progressively. The first query aggregates all agent features (dashed line) while the second only gathers features for every 2 time intervals (solid line). Subsequently, these features are fused into a single trajectory query, encompassing the hierarchical spatial-temporal features of the agent.}
   \label{fig:hqa}
\end{figure}

\paragraph{Context modelling.}
Trajectory prediction is inherently a sequential and interaction task, where an agent's past actions and the surrounding environment both play crucial roles. To extract spatial-temporal features of agents, we model map-map, agent-map, and agent-agent interactions sequentially.
Firstly, we model map structures and lane interactions, regarded as map priors, using the map-map attention module with an attention mechanism proposed in~\cite{transformer}, followed by a feed-forward network (FFN). The attention mechanism is defined as 
\begin{equation}
  Attention(Q,K,V) = softmax(\frac{QK^T}{\sqrt{d_k}})V,
  \label{eq:attn}
\end{equation}
where $Q$, $K$ and $V$ represent $Query$, $Key$ and $Value$ features, and $d_k$ denotes the feature dimension of $Key$. The interaction module is defined as %
\begin{multline}
      I(K, Q, V) = LN(f+FFN(f)) \\ 
  where \  f = LN(Q+Attention(Q,K,V)),
  \label{eq:attn-layer}
\end{multline}
where LN is the Layer Normalization~\cite{ln}. 
Following~\cite{mtr++,qcnet}, we calculate the relative information between map lanes, including positions and headings, as the position embedding, and add it to the $K$ and $V$ in each $Q$-$K$-$V$ pair. This relative position embedding idea is employed in all attention modules of our network. Second, agents incorporate map priors by interacting with nearby lane features in the agent-map attention module. Subsequently, to extract spatial and temporal features of agents efficiently, we use factorized attention which splits agent-agent interactions into temporal attention and spatial attention modules without sacrificing the performance as depicted in~\cite{wayformer}. For the agent-agent temporal attention, a lower triangular mask matrix is applied, ensuring that for any agent feature $F_i$, only features from previous time steps are used to compute attention. Relative time steps are also encoded as position embedding in this stage. For agent-agent spatial attention, only surrounding agents within a distance $d_n$ will be included. This process is repeated $L_E$ times to extract representative spatial-temporal agent features with map priors. Note that the map-free student model only contains agent-agent temporal and spatial attention modules.

\paragraph{Hierarchical feature aggregation and fusion.}
\label{graph:hqa}
Spatial-temporal agent features are aggregated using the Feature Aggregation (FA) module to generate multiple trajectory queries. These queries act as a bridge, connecting the encoder and decoder, and representing potential multimodal future trajectories. We achieve this by first initializing $K$ distinct trajectory queries and then performing cross-attention with spatial-temporal agent features. Specifically, we treat trajectory queries as $Query$ and the agent features as $Key$ and $Value$ when computing attention. This allows trajectory queries to absorb the spatial-temporal features of agents and be refined as the encoder progresses. To enhance the representation of future prediction modes with the trajectory queries, we explore hierarchical information embedded in trajectory points at various time intervals. We initialize $H$ different hierarchical queries, using them to aggregate agent features over time interval $2^{h-1}$, as depicted in \cref{fig:hqa}, where $h \in \{1,2,...,H\}$ is a hierarchical query index. After getting $H$ hierarchical queries $\{q_1,q_2,...,q_H\}$, we fuse them into a single trajectory query with %
\begin{equation}
  Q = MLP(cat(q_1,q_2,...,q_H)),
  \label{eq:hqa}
\end{equation}
where $cat$ denotes the concatenation operation. In conclusion, for each agent, we generate $KH$ hierarchical queries and obtain $K$ fused trajectory queries $\{Q^1,Q^2,...,Q^k\}$, corresponding to $K$ future trajectories during the decoding stage.

\subsection{Iterative Decoder}
\label{subsec:decoder}
In contrast to one-stroke methods~\cite{mtr,iccv2023adapt,wayformer} that output complete trajectories in a single pass or autoregressive methods~\cite{seff2023motionlm,li2022sit} that predict one step at a time, we employ the iterative batch decoding. This is achieved by predicting a fixed length of future (\eg 1 second) in each iteration, and the complete prediction is done in $I_T$ iterations. 
After getting $K$ trajectory queries, agents learn where to go by interacting with map priors via query-map attention. 
Then, we optimize the joint multimodal distribution of multiple future trajectories by applying self-attention within each agent. 
Following the query-query attention module, we employ simple MLPs to predict future trajectories as %
\begin{equation}
  S_i^k = MLP(Q_i^k),
  \label{eq:batch-pred}
\end{equation}
where $i \in \{1,2,...,N_A\}$, $k \in \{1,2,...,K\}$ and $S_i^k \in \mathbb{R}^{T' \times \{\Delta x, \Delta y, \sigma_x^2, \sigma_y^2\}}$. Here, $\Delta x$ and $\Delta y$ represent the relative motions between future points, which align with the input style and can recover absolute future positions with a simple summation operation. $\sigma_x^2$ and $\sigma_y^2$ denote the variances of Gaussian distribution. 
After completing $I_T$ iterations, we obtain the entire predicted trajectories by concatenating all the predictions $S = cat(S_i^k) \in \mathbb{R}^{N_A \times K \times N_T \times 4}$, where $N_T$ is the total future time steps, and a summation is performed over $\Delta x$ and $\Delta y$.

\subsection{Map Priors Distillation}
\label{subsec:distillation}
\paragraph{Teacher training.} The teacher network contains all components described above and is trained with the supervision of $L_{reg}$ and $L_{cls}$, as described in~\cref{subsec:loss}. Once trained, the teacher will keep fixed during the student distillation.

\paragraph{Knowledge transfer.}
We transfer map priors from the teacher to the student through intermediate feature distillation. This is possible because agent trajectories inherently reflect the topology of map lanes, allowing them to infer underlying map priors with guidance. We align the trajectory queries after the encoder and query features before the MLP in the decoder of the student with their counterpart features in the teacher. Specifically, given the trajectory queries for all $N_A$ agents $\{Q_j|j=1,2,...,N_AK\}$, we calculate the $L_2$ distance to align student features with teacher features as %
\begin{equation}
  L_{kd}^1 = \frac{1}{N_AK}(\sum_{j=1}^{N_AK} L_2(Q_j^s, Q_j^t)),
  \label{eq:kdq}
\end{equation}
where $Q_j^s$ is the $j$-th query feature of the student, $Q_j^t$ is the corresponding feature of the teacher. The minimization of the loss $L_{kd}^1$ ensures that the trajectory queries of the student learn from the teacher, thereby acquiring the map priors extracted during the encoding stage of the teacher.
To learn how to infer future motions with trajectory queries as if maps exist, we distill the query features before the MLP as
\begin{equation}
  L_{kd}^2 = \frac{1}{I_TN_AK}(\sum_{l=1}^{I_T}\sum_{j=1}^{N_AK} L_2(F_{l,j}^s, F_{l,j}^t)),
  \label{eq:kdd}
\end{equation}
where $F_{l,j}^s$ is the $j$-th student query feature in the $l$-th iteration and $F_{l,j}^t$ is the corresponding feature of the teacher.
Our final knowledge distillation loss is
\begin{equation}
  L_{kd} = L_{kd}^1 + L_{kd}^2,
  \label{eq:kd}
\end{equation}
which allows us to transfer map priors to our map-free student network on both the encoding and decoding stages.

\subsection{Training Objectives}
\label{subsec:loss}
In addition to the KD loss utilized in student training, we incorporate a regression loss for predicted trajectories and a classification loss for corresponding confidences. Following ~\cite{mtr,chai2019multipath,varadarajan2022multipath++}, we model predicted trajectories as a mixture of Gaussians. The optimal prediction trajectory for each agent is selected by finding the best match with the ground truth based on the average displacement error of all predicted points. We then treat predicted trajectories as the means of Gaussians and apply the negative log-likelihood loss as
\begin{multline}
  L_{reg} = -log(\frac{1}{2 \pi \sigma_x \sigma_y}\exp(-\frac{(x-\mu_x)^2}{2\sigma_x^2}-\frac{(y-\mu_y)^2}{2\sigma_y^2})) \\
  = log(2 \pi \sigma_x \sigma_y) + \frac{(x-\mu_x)^2}{2\sigma_x^2}+\frac{(y-\mu_y)^2}{2\sigma_y^2},
  \label{eq:guassian_nll}
\end{multline}
where $\mu_x$ and $\mu_y$ represent ground-truth future points. The final regression loss is obtained by first summing over all valid points of the matched trajectory with the ground truth for each agent and then averaging over all agents.
The classification loss is a simple cross-entropy loss that encourages the matched trajectory to have the highest confidence score as %
\begin{equation}
  L_{cls} = \frac{1}{N_A}(\sum_{i=1}^{N_A} CE(S_i, I_i)),
  \label{eq:cls_loss}
\end{equation}
where $S_i$ represents confidence scores of all $K$ predictions for agent $A_i$, and $I_i$ is the index of the matched prediction.
The total loss is defined as %
\begin{equation}
  Loss = \alpha L_{reg} + \beta L_{cls} + \gamma L_{kd},
  \label{eq:total_loss}
\end{equation}
where $\alpha$, $\beta$, and $\gamma$ are used to balance the losses for each component.

\begin{table*}[htbp]
  \centering
  \begin{tabular}{c l  c c c c c}
    \toprule[1pt]
       & Method & minADE$_6 \downarrow$ & minFDE$_6 \downarrow$ & MR$_6 \downarrow$ & brier-minFDE$_6 \downarrow$ & DAC$_6 \uparrow$\\
       \midrule
       & CRAT-Pred~\cite{cartpred}                          & 1.06 & 1.90 & 0.26 & -  & - \\
       & Fast map-free~\cite{fastmapfree}                   & 0.93 & 1.59 & 0.21 & -  & - \\
       & VectorNet$^\dagger$~\cite{gao2020vectornet}        & 1.16 & 2.22 & 0.37 & 2.91 & - \\
       & LaneGCN$^\dagger$~\cite{liang2020learning}         & 0.98 & 1.71 & 0.23 & 2.40 & - \\
   Map-free    & HiVT-64$^\dagger$~\cite{hivt}                      & 0.95 & 1.64 & 0.22 & 2.32  & - \\
       & HiVT-128$^\dagger$~\cite{hivt}                     & 0.91 & 1.54 & 0.20 & 2.21 & - \\
       & HiVT-128$^\dagger$ + FOKD~\cite{wang2023enhancing} & 0.88 & 1.47 & \underline{0.18} & 2.15 & - \\
       & MFTP-NKD (ours)                                    & \underline{0.86} & \underline{1.43} & \underline{0.18} & \underline{2.10}  & \underline{0.94} \\
       & MFTP (ours)                                        & \textbf{0.84}    & \textbf{1.38}    & \textbf{0.16}    & \textbf{2.03}  & \textbf{0.96} \\
        
       \midrule

       & HiVT-64$^{*}$~\cite{hivt}       & 0.83 & 1.31 & 0.15 &  1.97  & - \\
   Map-based    & HiVT-128$^{*}$~\cite{hivt}      & 0.80 & 1.23 & 0.14 &  1.90  & - \\
       & MFTP-T (ours)                   & 0.80 & 1.22 & 0.13 &  1.85  & 0.99 \\
       & MFTP-T$^\ddagger$ (ours)        & 1.90 & 3.77 & 0.56 &  4.48  & 0.87 \\
       
  \bottomrule[1pt]
  \end{tabular}
  \caption{\textbf{Prediction results on Argoverse test set.} $\dagger$: Map-based method rerun under map-free setting, and we get the map-free performance of these methods from ~\cite{wang2023enhancing}. $^{*}$: Performance reproduced by ~\cite{wang2023enhancing}.
  MFTP-NKD represents the student network without distillation, while MFTP-T denotes the teacher network. MFTP-T$^\ddagger$ is teacher network but test without map.
  For each metric, the best result is in \textbf{bold} and the second best is \underline{underlined}.}
  \label{tab:argoverse-test}
\end{table*}

\begin{table}[htbp]
  \centering
  \resizebox{0.99\linewidth}{!}{
  \begin{tabular}{l  c c c}
    \toprule[1pt]
       Method & minADE$_6 \downarrow$ & minFDE$_6 \downarrow$ & MR$_6 \downarrow$\\
    \midrule

        CRAT-Pred     & 0.85 & 1.44 & 0.17 \\
       Fast map-free & 0.74 & 1.18 & 0.12 \\
        VectorNet$^\dagger$  & 0.93 & 1.70 & 0.24 \\
        LaneGCN$^\dagger$  & 0.79 & 1.29 & 0.15 \\
 HiVT-64$^\dagger$  & 0.77 & 1.25 & 0.14 \\
       HiVT-128$^\dagger$  & 0.73 & 1.15 & 0.13 \\
       HiVT-128$^\dagger$ + FOKD  & 0.71 & 1.11 & \textbf{0.11}\\
       MFTP-NKD (ours) & \underline{0.70} & \underline{1.10} & \underline{0.12}\\
       MFTP (ours) & \textbf{0.68} & \textbf{1.07} & \textbf{0.11}\\
    \midrule
       
       HiVT-64$^{*}$        & 0.69 & 1.04 & 0.10\\
       HiVT-128$^{*}$       & 0.66 & 0.97 & 0.09\\
       MFTP-T (ours) & 0.65 & 1.01 & 0.09\\

  \bottomrule[1pt]
  \end{tabular}
  }
  \caption{\textbf{Prediction results on Argoverse validation set.} The table setting is the same as \cref{tab:argoverse-test}.}
  \label{tab:argoverse-val}
\end{table}

\section{Experiments}
\label{sec:exp}
In this section, we describe the experimental setup used to evaluate our method, and the evaluation results.

\subsection{Experimental Setup}
\paragraph{Dataset}
We evaluate our method on the large-scale motion forecasting dataset, Argoverse~\cite{argoverse1}. Argoverse contains 333441 high-quality five-second sequences for motion forecasting, covering diverse scenarios such as (1) at intersections (2) taking left or right turns (3) changing to adjacent lanes, or (4) in dense traffic. Among these scenarios, 63.5\% are partitioned for training, 12.3\% for validation, and 24.2\% for testing. 
Each agent in the dataset has a history of 2 seconds and a future trajectory of 3 seconds, sampled at 10 Hz.

\paragraph{Metrics}
We follow the official metrics used in Argoverse, including Minimum Final Displacement Error (minFDE), Minimum Average Displacement Error (minADE), Miss Rate (MR), Brier minimum Final Displacement Error (brier-minFDE) and Drivable Area Compliance (DAC). 
minFDE measures the L2 distance of the endpoint, whereas minADE calculates the average L2 distance of all valid points. MR evaluates the validity of predictions by flagging a miss if none of the prediction trajectories for an agent falls within a circle centered at the ground-truth endpoint. Besides the prediction accuracy, prediction confidence is also important for the subsequent planning task. To punish the low-confidence prediction, brier-minFDE adds a score item $(1-p)^2$ to minFDE, where $p$ is the probability of the matched trajectory. DAC measures the ratio of predicted trajectories within drivable areas.

\begin{table*}[htbp]
    \centering
    \begin{tabular}{cccc|cccc}
    \toprule[1pt]
        AATA & AASA & FA & QQA & minADE$_6 \downarrow$ & minFDE$_6 \downarrow$ & MR$_6 \downarrow$ & brier-minFDE$_6 \downarrow$ \\
    \midrule
       \xmark  & \cmark & \cmark & \cmark & 0.81 & 1.37 & 0.16 & 2.03 \\ 
       \cmark  & \xmark & \cmark & \cmark & 0.89 & 1.62 & 0.22 & 2.28 \\ 
       \cmark  & \cmark & \xmark & \cmark & 0.82 & 1.38 & 0.16 & 2.04 \\ 
       \cmark  & \cmark & \cmark & \xmark & 0.86 & 1.31 & \textbf{0.15} & 1.96 \\ 
       \cmark  & \cmark & \cmark & \cmark & \textbf{0.78} & \textbf{1.30} & \textbf{0.15} & \textbf{1.95} \\ 
    \bottomrule[1pt]
    \end{tabular}
    \caption{\textbf{Ablation studies on proposed attention modules.} AATA: agent-agent temporal attention. AASA: agent-agent spatial attention. FA: features aggregation. By removing FA, we instead use the most recent feature of agents and append $K$ randomly initialized mode features to form $K$ trajectory queries used in the decoder. QQA: query-query attention. All results are obtained with our map-free model, MFTP-NKD. The best result is in \textbf{bold} for each metric.}
    \label{tab:abalation-attention}
\end{table*}

\begin{table}[htbp]
    \centering
    \begin{tabular}{c | c c c}
    \toprule[1pt]
       $H$  & minADE$_6 \downarrow$ & minFDE$_6 \downarrow$ & MR$_6 \downarrow$  \\
         \midrule
       1  & 0.80          & 1.33          & 0.16 \\
       2  & 0.79          & 1.31          & \textbf{0.15} \\
       3  & \textbf{0.78} & \textbf{1.30} & \textbf{0.15} \\
       4  & \textbf{0.78} & \textbf{1.30} & \textbf{0.15} \\
    \bottomrule[1pt]
    \end{tabular}
    \caption{\textbf{Ablation studies of the number of hierarchical queries.} All results are obtained with our map-free model, MFTP-NKD. The best result is in \textbf{bold} for each metric.}
    \label{tab:ablation-hq}
\end{table}

\paragraph{Implementation Details}
For the hierarchical encoder, we stack $L_E = 3$ layers and initialize $N_AKH$ hierarchical queries, where $K=6$ and $H=3$, to extract hierarchical features for all $N_A$ agents. Note that Argoverse also requires predicting 6 future trajectories, therefore, we avoid the need to select top-$K$ prediction with any post-processing techniques used in other methods (\eg NMS~\cite{mtr}, K-means~\cite{seff2023motionlm}). For the iterative decoder, we set $L_D=3$, where each iteration of the MLP predicts 1 second of future trajectories. Therefore, we set $I_T=3$ for Argoverse, predicting trajectories for the next 3 seconds. For the spatial attention in map-map, agent-map, and agent-agent spatial attention modules, we consider only neighbors within a 100-meter radius. We first train the teacher model with the supervisions of $L_{reg}$ and $L_{cls}$ and keep the model fixed when training the student model with the supervisions of $L_{reg}$, $L_{cls}$ and $L_{KD}$. The loss weights $\alpha = 1$, $\beta = 1$, and $\gamma = 1$ are set for all experiments. We use the Adam optimizer, setting the initial learning rate to $1 \times 10^{-4}$ and incorporating a cosine annealing scheduler for 32 epochs.

\begin{figure*}[t]
  \centering
  \begin{subfigure}{1\linewidth}
    \includegraphics[width=0.99\textwidth]{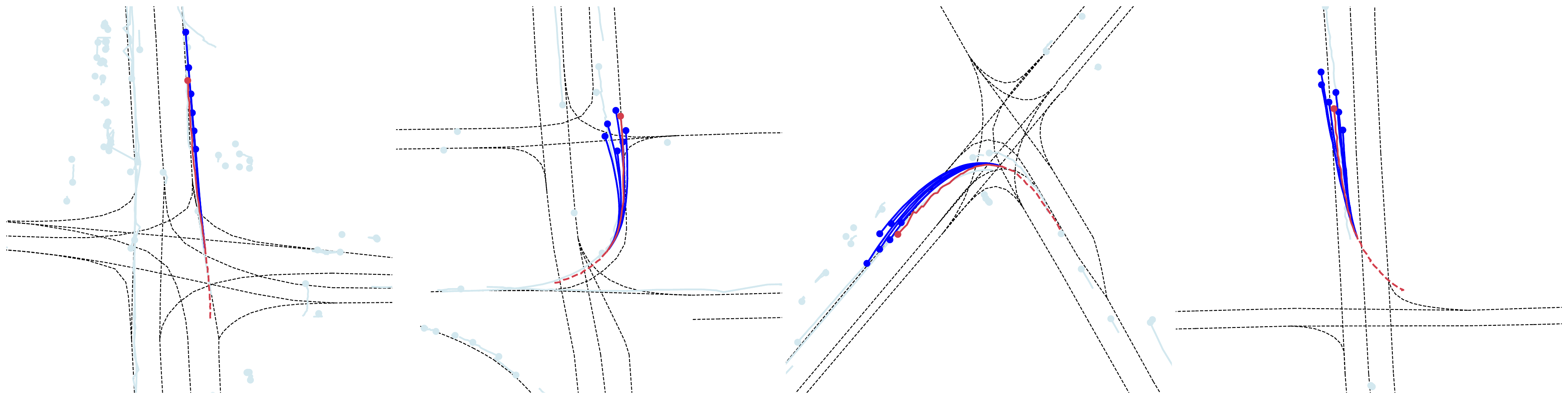}
    \caption{Map-free prediction without KD}
    \label{fig:nkd}
  \end{subfigure}
  
  \vfill

  \begin{subfigure}{1\linewidth}
    \includegraphics[width=0.99\textwidth]{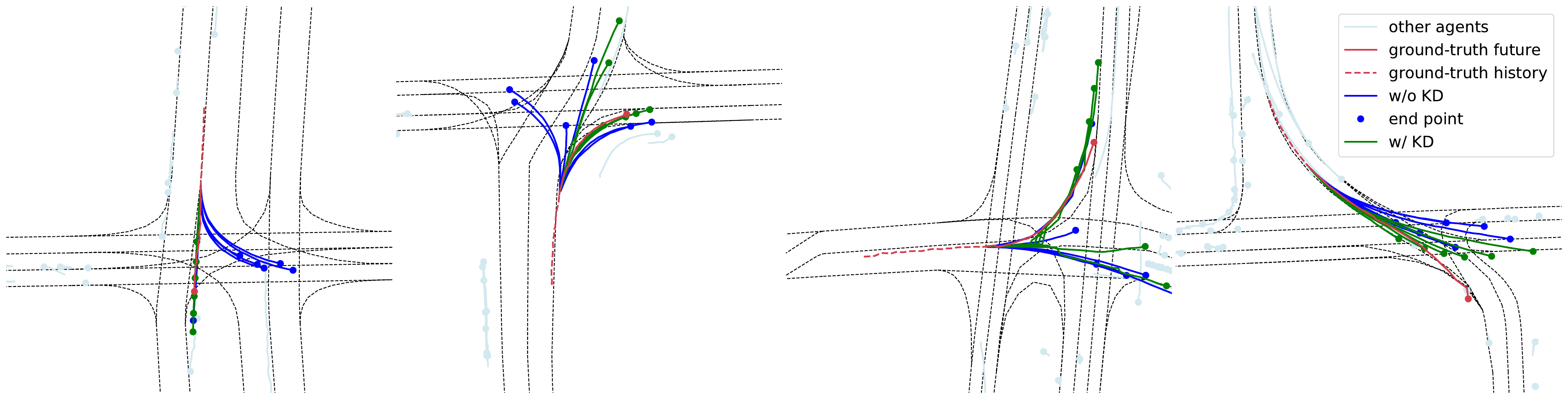}
    \caption{Map-free prediction with KD}
    \label{fig:kd}
  \end{subfigure}

   \caption{\textbf{Qualitative results on Argoverse validation set.} (a) illustrates the performance of our map-free model on intersection scenarios with various driving behaviors (\eg, go-straight, left-turn, big left-turn and right-turn from left to right) without leveraging map priors. (b) demonstrates that, with the help of knowledge distillation (KD), our map-free model can predict future trajectories more closely aligned with the ground truth. (a) and (b) share the same figure legend. Best viewed in color and zoomed in.}
   \label{fig:vis}
\end{figure*}

\subsection{Quantitative Results}
We compare our method with state-of-the-art map-free methods on Argoverse, as shown in \cref{tab:argoverse-test} and \cref{tab:argoverse-val}. Without knowledge distillation, our map-free method, MFTP-NKD, outperforms all other non-distilled map-free methods across all metrics on both validation and test sets. Notably, MFTP-NKD also surpasses the knowledge-distilled HiVT-128$^\dagger$ + FOKD in most metrics, which proves the effectiveness of our proposed map-free framework. After applying the knowledge distillation, we further improve our model's performance, enlarging the performance margin over other methods. The improvement of DAC$_6$ shows the student model successfully learned map priors through knowledge distillation. These comparative results demonstrate the superiority of our overall design for map-free prediction. Additionally, the necessity of a map-free model is further highlighted by MFTP-T$^\ddagger$'s performance in \cref{tab:argoverse-test}, as map-based model degrade significantly without taking maps as input.

\begin{table}[htbp]
    \centering
    \begin{tabular}{c c | c c c}
    \toprule[1pt]
       \#iter &  length  & minADE$_6 \downarrow$ & minFDE$_6 \downarrow$ & MR$_6 \downarrow$  \\
         \midrule
       1  & 3s   & 0.81          & 1.34          & 0.17 \\
       2  & 1.5s & 0.79          & 1.32          & 0.16 \\
       3  & 1s   & \textbf{0.78} & 1.30          & \textbf{0.15} \\
       6  & 0.5s & \textbf{0.78} & \textbf{1.29} & \textbf{0.15} \\
    \bottomrule[1pt]
    \end{tabular}
    \caption{\textbf{Ablation studies of iterative decoding steps.} We predict 3-second future trajectories as required by Argoverse. All results are obtained with our map-free model, MFTP-NKD. The best result is in \textbf{bold} for each metric.}
    \label{tab:iter}
\end{table}

\subsection{Ablation Studies}
All ablation results are obtained with our map-free model, MFTP-NKD, without KD to verify the effectiveness of proposed attention modules, hierarchical query aggregation, and iterative decoder. We train the models with 20$\%$ training data and evaluate on full validation set.

\paragraph{Effects of attention modules.}
We study the effectiveness of our proposed attention modules, namely AATA, AASA, FA, and QQA. As shown in \cref{tab:abalation-attention}, all attention modules significantly contribute to the final results. Notably, the absence of AATA leads to a 5.4\% increase in minFDE$_6$, highlighting the importance of leveraging historical temporal information to enhance future prediction accuracy. Similar results are observed for the FA module, which extracts hierarchical features at temporal level, complementing the role of the AATA module. Moreover, the exclusion of AASA results in a significant performance decline, with minADE$_6$ increasing by 14.1\%, minFDE$_6$ by 24.6\%, MR$_6$ by 46.7\%, and brier-minFDE$_6$ by 16.9\%. This ablation result underscores the crucial role of spatial relationships in map-free trajectory prediction. It supports our hypothesis that agents generally follow the map lanes, and their trajectories inherently reflect the map structure. It also confirms that the map-free student model can learn map priors through knowledge distillation, even with only agent historical trajectories. Finally, the removal of the QQA module leads to a notable 10.3\% increase in minADE$_6$, emphasizing the importance of the interactions between multimodal queries. The ablation study collectively highlights the indispensability of each attention module in our proposed framework.

\paragraph{Effects of the number of hierarchical queries.}
As described in ~\cref{graph:hqa}, we initialize $KH$ hierarchical queries for each agent and ultimately get $K$ trajectory queries. Here, we present ablation studies on different values of $H$ in \cref{tab:ablation-hq}. As the number of $H$ increases, we observe an improvement in distance-based metrics, namely minADE$_6$ and minFDE$_6$. However, we also notice that the performance increase stops when $H=4$. This is because, for the fourth hierarchical query, only 2 historical features are extracted given the 20 historical time steps, as the query looks at features with $2^{H-1}=8$ time intervals. Therefore, we use $H=3$ in our final model.

\paragraph{Effects of iteration steps of the decoder.}
We adopt an iterative decoding style to balance accuracy and efficiency. \cref{tab:iter} presents the ablation results for different prediction lengths. With 1 iteration, we predict the entire future trajectory in one pass, yielding the worst results for our model. As we reduce the prediction length, we observe an improvement in performance.  However, the improvement gradually decreases as the number of iterations increases. Despite the fact that 6 iterations result in a better minFDE$_6$ compared to 3 iterations, we choose to set the model to predict 1 second of the future at a time for efficiency.

\subsection{Qualitative Results}
We present qualitative results on the Argoverse validation set in \cref{fig:vis}. In \cref{fig:nkd}, our map-free model shows the ability to predict accurate future trajectories purely based on agents' historical trajectories. This success can be attributed to interactions with other agents, as their trajectories provide valuable guidance for predicting the trajectories of interested agents. In \cref{fig:kd}, the predictions of our map-free model are not well-aligned with the ground truth. However, after applying the knowledge distillation, which provides additional hints of map priors, our map-free model achieves better results in these challenging scenarios.

\section{Conclusion}
\label{sec:conlcusion}
In this paper, we introduce MFTP, a map-free trajectory prediction method that incorporates map priors through knowledge distillation. The hierarchical encoder effectively captures multi-level spatial-temporal agent features, which are then fused into multiple trajectory queries.
These queries are subsequently employed to predict multimodal future trajectories iteratively with our decoder. The results show that our method attains state-of-the-art performance on the Argoverse dataset under the map-free setting.

{\small
\bibliographystyle{ieee_fullname}
\bibliography{egbib}
}

\end{document}